\documentclass[conference]{IEEEtran}
\IEEEoverridecommandlockouts
\usepackage{cite}
\usepackage{amsmath,amssymb,amsfonts}
\usepackage{algorithmic}
\usepackage[colorlinks=true, urlcolor=blue, linkcolor=blue, citecolor=blue]{hyperref}
\usepackage{graphicx}
\usepackage{textcomp}
\usepackage{xcolor}
\usepackage{booktabs}
\def\BibTeX{{\rm B\kern-.05em{\sc i\kern-.025em b}\kern-.08em
    T\kern-.1667em\lower.7ex\hbox{E}\kern-.125emX}}
\begin{document}


\title{Early Prediction of Type 2 Diabetes Using Multimodal data and Tabular Transformers\\

\thanks{The publication of this article was funded by Qatar National Library.}
}

\author{\IEEEauthorblockN{Sulaiman Khan, Md. Rafiul Biswas, Zubair Shah\textsuperscript{*}}
\IEEEauthorblockA{\textit{College of Science and Engineering, Hamad Bin Khalifa University, Qatar Foundation, Doha, Qatar} \\ \\
sukh45452@hbku.edu.qa; mbiswas@hbku.edu.qa; zshah@hbku.edu.qa}

}

\maketitle

\begin{abstract}
This study introduces a novel approach for early Type 2 Diabetes Mellitus (T2DM) risk prediction using a tabular transformer (TabTrans) architecture to analyze longitudinal patient data. By processing patients longitudinal health records and bone-related tabular data, our model captures complex, long-range dependencies in disease progression that conventional methods often overlook.
We validated our TabTrans model on a retrospective Qatar BioBank (QBB) cohort of 1,382 subjects, comprising 725 men (146 diabetic, 579 healthy) and 657 women (133 diabetic, 524 healthy). The study integrated electronic health records (EHR) with dual-energy X-ray absorptiometry (DXA) data. To address class imbalance, we employed SMOTE and SMOTE-ENN resampling techniques.
The proposed model's performance is evaluated against conventional machine learning (ML) and generative AI models, including Claude 3.5 Sonnet (Anthropic's constitutional AI), GPT-4 (OpenAI's generative pre-trained transformer), and Gemini Pro (Google's multimodal language model). Our TabTrans model demonstrated superior predictive performance, achieving ROC AUC $\ge$ 79.7\% for T2DM prediction compared to both generative AI models and conventional ML approaches.
Feature interpretation analysis identified key risk indicators, with visceral adipose tissue (VAT) mass and volume, ward bone mineral density (BMD) and bone mineral content (BMC), T and Z-scores, and L1-L4 scores emerging as the most important predictors associated with diabetes development in Qatari adults.
These findings demonstrate the significant potential of TabTrans for analyzing complex tabular healthcare data, providing a powerful tool for proactive T2DM management and personalized clinical interventions in the Qatari population.

\end{abstract}


\section{Introduction}
Longitudinal analysis of bone-related measurements has recently emerged as an important tool in exploring the risk factors associated with type 2 diabetes mellitus (T2DM). Several studies have shown that bone mineral density (BMD) and bone mineral content (BMC), measured through dual-energy X-ray absorptiometry (DXA), are not only indicators of skeletal health but are also associated with metabolic disorders including diabetes \cite{gan2017bone, leslie2018comparison}. Moreover, longitudinal DXA data from biobank cohorts have revealed significant associations between bone composition changes and the future development of T2DM, highlighting the potential of skeletal imaging as a predictive biomarker \cite{khan2025artificial}.

Globally, T2DM is followed as a major health concern that directly affect nearly half a billion people worldwide, accurate and early prediction remains critical for reducing disease burden \cite{tabak2012prediabetes, federation2025idf}. With the availability of large-scale biobank datasets, such as the Qatar Biobank (QBB), it is now possible to study diabetes risk factors using rich longitudinal imaging and clinical data. However, traditional statistical models and conventional machine learning methods often struggle to capture the complex temporal and spatial patterns inherent in biomedical imaging data. This necessitates the adoption of advanced deep learning architectures that can efficiently model these high-dimensional relationships.

Recent work has moved beyond tabular features to exploit multimodal data—combining electronic health records (EHRs), imaging, metabolomics, and demographic information—to improve prediction accuracy. For example, metabolomics integrated with machine learning models has enhanced early T2DM prediction \cite{ronn2024predicting, kavakiotis2017machine}, while multimodal frameworks using chest X-ray images and EHR data have shown promise for automated diabetes screening \cite{choi2024novel}. At the same time, the emergence of transformer architectures has reshaped predictive modeling. Originally introduced in natural language processing, transformers have been adapted for tabular data, demonstrating strong performance in capturing feature interactions and handling missing or noisy data \cite{kadra2021well, huang2020tabtransformer}. Moreover, multimodal transformer frameworks have been successfully applied in healthcare to integrate heterogeneous data sources, suggesting their potential in chronic disease prediction \cite{kumar2025transformer}.

Similarly, vision transformers (ViTs) have gained prominence in medical imaging tasks by leveraging self-attention mechanisms to capture global dependencies across image regions \cite{khan2022transformers, khan2025diabeye}. Unlike convolution neural networks (CNNs), which rely primarily on local receptive fields, ViTs can model long-range spatial interactions, making them particularly well-suited for multimodal and longitudinal biomedical data. Applications of ViTs in healthcare, such as diabetic retinopathy classification from retinal fundus images, have demonstrated superior performance compared to traditional CNN-based approaches \cite{mutawa2022diabetic, lin2023development}. Similarly, Sumon et al., \cite{sumon2025cardiotabnet} proposed a transformer-based model named CardioTabNet for cardiovascular disease prediction using tabular clinical data including cholesterol levels information, fasting blood sugar, and others.

Despite these advances, limited research has been conducted on applying tabular transformer (TabTrans) to longitudinal imaging data (bone data) for T2DM prediction. Moreover, the longitudinal investigations of bones parametric data versus diabetes development with time in this context remains underexplored. This motivates the present study, where we propose a tabular transformer-based framework that utilizes longitudinal bone measurements from QBB to predict diabetes risk. Specifically, our main contributions are: (i) to generate features from DXA data and use use TabTrans for annotating and interpreting longitudinal bone data, (ii) to perform feature interpretation for identifying risk factors associated with diabetes development in Qatari adults, and (iii) to perform a comparative analysis of TabTrans against typical machine learning models and generative AI models including Claude 3.5 Sonnet (Anthropic's constitutional AI model), GPT-4 (OpenAI's generative pre-trained transformer), and Gemini Pro (Google's multimodal language model).

Our analysis revealed that TabTrans demonstrated superior performance compared to generative AI models but showed lower performance than conventional statistical machine learning (ML) approaches. These findings suggest that for tabular datasets with limited sample sizes, traditional ML algorithms such as XGBoost, LightGBM, and random forest represent more effective alternatives. The conventional ML methods achieved higher precision scores and AUC values compared to both TabTrans and generative AI models evaluated in this study. 

\section{Methodology}
\begin{figure}
    \centering
    \includegraphics[width=1\linewidth]{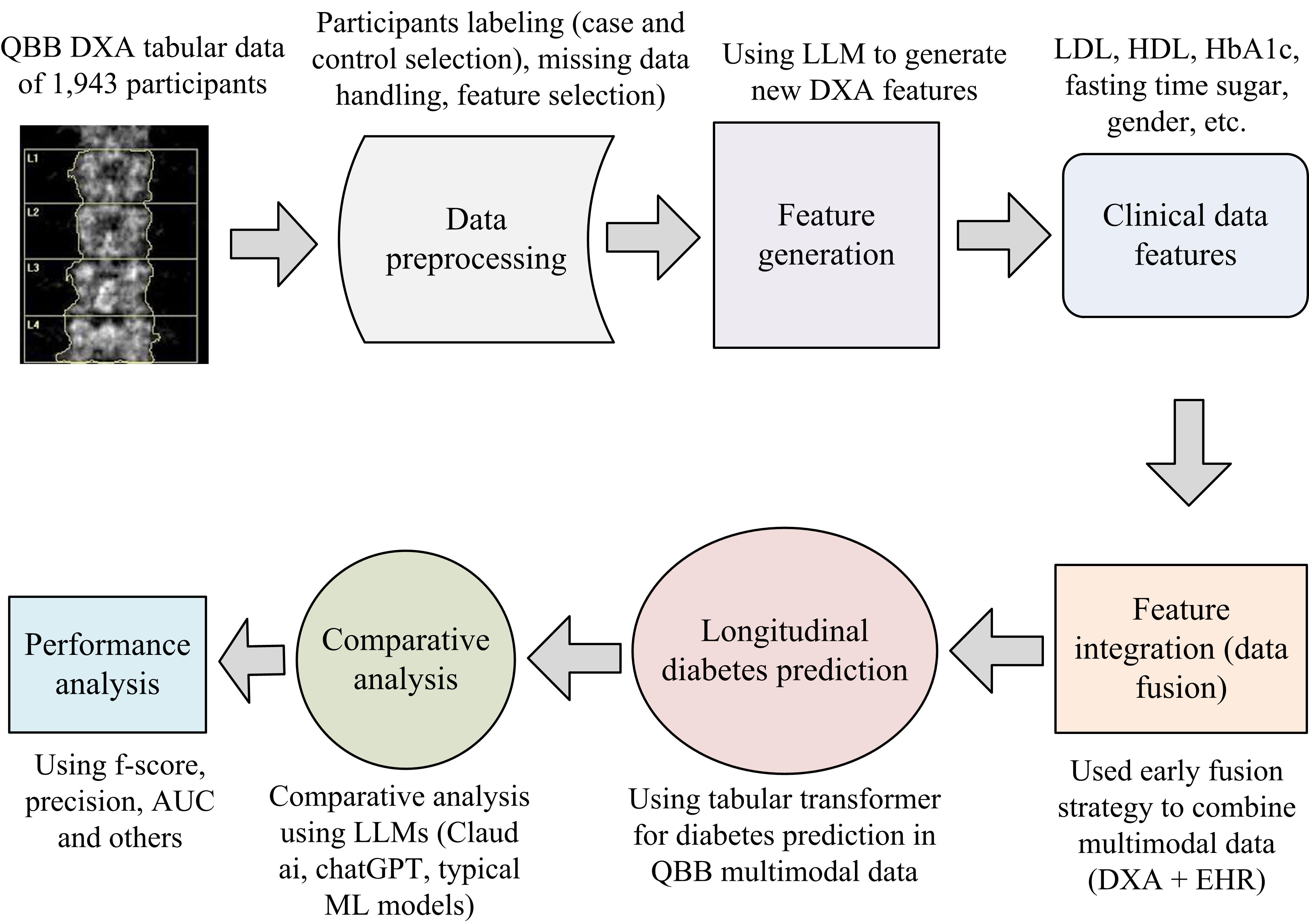}
    \caption{Systematic methodology followed to perform this research work.}
    \label{fig:process_diag}
\end{figure}

A systematic methodology was implemented to conduct this longitudinal diabetes investigation utilizing RDS information, as illustrated in Fig. \ref{fig:process_diag}. Sequential phases were executed to complete this research endeavor, encompassing data preparation and labeling, feature derivation and model development, predictive analysis and result interpretation. Each of these sequential components is detailed in the following sections. It is worth noting that for this simulation we used a simple PC with 8GB RAM and Intel Processor Core-i7 with no GPU installed.

\subsection{Ethics Committee Authorization}
This investigation operated under supervision from Qatar's Ministry of Public Health (MOPH) and obtained ethics approval from the institutional review committees of both QBB and Hamad Bin Khalifa University. The study utilized de-identified information supplied by QBB while maintaining compliance with established protocols and regulatory standards. QBB confirmed that appropriate informed consent documentation had been obtained from all research participants and/or their authorized representatives.

\subsection{Data Collection and Preprocessing}
This study utilizes multimodal data from the Qatar Biobank (QBB), collected through standardized acquisition protocols \cite{al2015qatar, al2019qatar}. The dataset comprises diverse modalities, including genetic, DXA, clinical, and retinal fundoscopic data, among others. In this work, we focus on integrating clinical and DXA data through feature engineering to predict diabetes onset using the TabTrans model. The QBB dataset includes two longitudinal components: baseline and follow-up. The baseline dataset captures participants’ initial clinical assessments, whereas the follow-up dataset contains longitudinal records from subsequent visits over time. Following criteria is used to classify participants as healthy or diabetic, removing missing values.

\begin{itemize}
    \item \textbf{Diabetic criteria \textemdash} Self-reported T2DM OR HbA1c $\ge$ 7\% OR fasting glucose $\ge$ 126 mg/dL OR current use of antidiabetic medications/insulin therapy.

    \item \textbf{Preprocessing \textemdash} The DXA variables containing more than 50\% missing observations were eliminated. Additionally, forward-fill imputation was applied to address remaining null values in the dataset.
\end{itemize}

This research specifically investigates T2DM among individuals aged 25–84 years. From the initial cohort of 1,941 participants, 1,382 individuals had available follow-up DXA scan data. The study population contains: T2DM cases (n=287) and controls (n=1,095). It comprising 725 males with (579 control, 146 cases) and 657 females with (524 control, 133 cases). To address class imbalance and augment this tabular data before training our TabTrans model, we applied the following steps:

\begin{itemize}
    \item \textbf{Class imbalance \textemdash} To address class imbalance, an ensemble of advanced resampling techniques—including Borderline-SMOTE, SVM-SMOTE, ADASYN, SMOTE-Tomek, and SMOTE-ENN—was evaluated. The optimal method was selected based on achieving a balanced class distribution while preserving data integrity, with adaptive k-nearest neighbor parameters automatically tuned according to minority class size.

    \item \textbf{Minority Class Augmentation \textemdash} A comprehensive augmentation pipeline was implemented to enhance minority class representation, incorporating noise injection ($\sigma = 0.2$), mixup augmentation ($\alpha = 0.6$), and weighted random sampling with a $10{:}1$ minority-to-majority ratio. An augmentation factor of $5\times$ was applied to the minority class to ensure sufficient representation during training.
\end{itemize}




\subsection{Problem Modeling}
The problem in this study was modeled to predict the future onset of T2DM. To achieve this, follow-up records were analyzed to identify individuals who developed diabetes over time, and these outcomes were used as labels for the corresponding baseline data. This linking of baseline features with future outcomes allows the model to learn patterns that precede disease onset. The labeled dataset was divided into training and testing subsets using stratified sampling. The training data were used to develop and optimize the predictive models, while the testing data evaluated generalization performance. Three modeling paradigms were explored: (i) conventional machine learning algorithms as baselines, (ii) large language models prompted for structured prediction, and (iii) the proposed TabTrans architecture, which employs self-attention to capture complex dependencies within tabular healthcare data. Through this formulation, the model learns associations between DXA-derived and clinical features that reflect early metabolic and skeletal changes linked to diabetes development.
\subsection{Feature Engineering and Selection}
This section describes the mathematical foundations underlying feature engineering and selection methodologies for diabetes risk prediction using DXA measurements. For DXA data, we generated features and for clinical data we selected features based on correlation analysis and features importance. Finally, we combined both DXA features and clinical features using early fusion strategy to create an optimized framework for clinical interpretability and predictive performance.

\subsubsection{Feature Generations} 
Our clinical DXA diabetes prediction system implements a focused set of feature engineering methods based on established diabetes pathophysiology. Ten Implemented Engineered Features are:

\subsubsection*{Central Obesity Ratio}
The primary diabetes risk indicator implemented:

\begin{equation}
\text{Central\_Obesity\_Ratio} = \frac{\text{Android Fat Mass}}{\text{Gynoid Fat Mass} + \epsilon}
\label{eq:central_obesity}
\end{equation}

where $\epsilon = 10^{-8}$ prevents division by zero.
\subsubsection*{Visceral Adiposity Index}
Quantifies visceral fat distribution:

\begin{equation}
\text{Visceral\_Adiposity\_Index} = \frac{\text{VAT Mass}}{\text{Total Fat Mass} + \epsilon}
\label{eq:visceral_adiposity}
\end{equation}

\subsubsection*{Muscle-Fat Ratio}
Sarcopenia and insulin resistance indicator:

\begin{equation}
\text{Muscle\_Fat\_Ratio} = \frac{\text{Total Lean Mass}}{\text{Total Fat Mass} + \epsilon}
\label{eq:muscle_fat_ratio}
\end{equation}

\subsubsection*{Trunk Fat Percentage}
Central adiposity distribution:

\begin{equation}
\text{Trunk\_Fat\_Percentage} = \frac{\text{Trunk Fat Mass}}{\text{Total Fat Mass} + \epsilon} \times 100
\label{eq:trunk_fat_percentage}
\end{equation}

\subsubsection*{Fat-Free Mass Index}
Area-adjusted lean mass:

\begin{equation}
\text{FFM\_Index} = \frac{\text{Total Fat-Free Mass}}{\sqrt{\text{Total Area} + \epsilon}}
\label{eq:ffm_index}
\end{equation}

\subsubsection*{Mean Spine BMD}
Average spine bone mineral density:

\begin{equation}
\text{Spine\_BMD\_Mean} = \frac{1}{n} \sum_{i \in \{\text{L1,L2,L3,L4}\}} \text{BMD}_i
\label{eq:spine_bmd_mean}
\end{equation}

where $n$ is the number of available spine BMD measurements.

\subsubsection*{Bone Health Composite}
Average T-score across all sites:

\begin{equation}
\text{Bone\_Health\_Composite} = \frac{1}{m} \sum_{j=1}^{m} \text{T-Score}_j
\label{eq:bone_health_composite}
\end{equation}

where $m$ is the number of available T-score measurements.

\subsubsection*{Osteoporosis Risk Score}
Count of T-scores indicating osteoporosis:

\begin{equation}
\text{Osteoporosis\_Risk} = \sum_{j=1}^{m} \mathbf{1}(\text{T-Score}_j < -2.5)
\label{eq:osteoporosis_risk}
\end{equation}

where $\mathbf{1}(\cdot)$ is the indicator function.

\subsubsection*{Peripheral Fat Ratio}
Fat distribution pattern:

\begin{equation}
\text{Peripheral\_Fat\_Ratio} = \frac{\text{Arms Fat Mass} + \text{Legs Fat Mass}}{\text{Trunk Fat Mass} + \epsilon}
\label{eq:peripheral_central_fat}
\end{equation}

\subsubsection*{BMD Coefficient of Variation}
Bone metabolism heterogeneity:

\begin{equation}
\text{BMD\_Coefficient\_Variation} = \frac{\sigma_{\text{BMD}}}{\mu_{\text{BMD}} + \epsilon}
\label{eq:bmd_cv}
\end{equation}

where $\sigma_{\text{BMD}}$ and $\mu_{\text{BMD}}$ are the standard deviation and mean of available BMD measurements.

\subsubsection{Dimensionality Reduction} 
\begin{figure*}[!ht]
    \centering
    \includegraphics[width=0.90\linewidth]{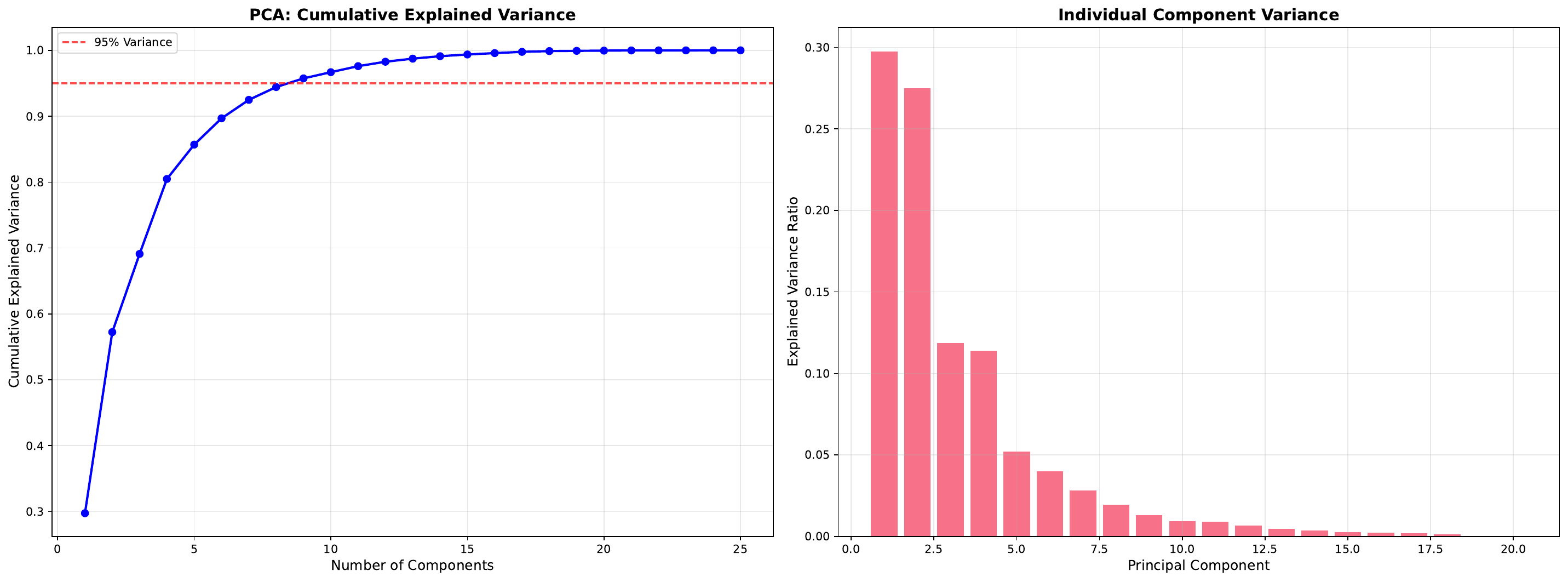}
    \caption{Feature reduction using PCA.}
    \label{fig:reduction_pca}
\end{figure*}

To explore the structure and separability of DXA-derived features, dimensionality reduction was conducted using PCA and t-SNE. PCA \cite{abdi2010principal} was applied to standardized features to retain 95\% of variance and visualize feature contributions via biplots (Fig.~\ref{fig:reduction_pca}). For datasets with fewer than 5,000 samples, t-SNE \cite{cieslak2020t} was used to visualize clustering between diabetic and non-diabetic groups, preceded by PCA reduction to 50 components when feature dimensionality exceeded this threshold for computational efficiency.

\subsubsection{Features Selection} 
Features were selected from clinical data using the following three main strategies including clinical correlation interpretation, BMD coefficient of variation, Pearson Correlation, and ensemble selection. In our case ensemble selection outperformed the other methodologies. 

\textbf{Clinical correlation interpretation \textemdash}
We establish clinical significance thresholds for correlation magnitudes using the constrints defined in Tab. ~\ref{tab:correlation_interpretation}.

\begin{table}[htbp]
\centering
\caption{Clinical Interpretation of Correlation Magnitudes}
\label{tab:correlation_interpretation}
\begin{tabular}{ll}
\toprule
Correlation Magnitude & Clinical Interpretation \\
\midrule
$|r| \geq 0.8$ & Very strong clinical association \\
$0.6 \leq |r| < 0.8$ & Strong clinical association \\
$0.4 \leq |r| < 0.6$ & Moderate clinical association \\
$0.2 \leq |r| < 0.4$ & Weak clinical association \\
$|r| < 0.2$ & Negligible clinical association \\
\bottomrule
\end{tabular}
\end{table}

\textbf{BMD Coefficient of Variation \textemdash}
Bone metabolism heterogeneity is calculated using Eq: (\ref{eq:bmd_cv}), while linear association with diabetes is calculated using Pearson correlation as shown in Eq: (\ref{eq:pearson_correlation}). A features selected if $|r_j| \geq 0.12$.

\begin{equation}
\text{BMD\_Coefficient\_Variation} = \frac{\sigma_{\text{BMD}}}{\mu_{\text{BMD}} + \epsilon}
\label{eq:bmd_cv}
\end{equation}

where $\sigma_{\text{BMD}}$ and $\mu_{\text{BMD}}$ are the standard deviation and mean of available BMD measurements.

\begin{equation}
r_{j} = \frac{\sum_{i=1}^{n} (X_{ij} - \bar{X}_j)(Y_i - \bar{Y})}{\sqrt{\sum_{i=1}^{n} (X_{ij} - \bar{X}_j)^2 \sum_{i=1}^{n} (Y_i - \bar{Y})^2}}
\label{eq:pearson_correlation}
\end{equation}

\begin{equation}
\text{MI}(X_j, Y) = \sum_{x \in \mathcal{X}_j} \sum_{y \in \mathcal{Y}} p(x,y) \log_2 \frac{p(x,y)}{p(x)p(y)}
\label{eq:mutual_information}
\end{equation}

\begin{figure*}[!ht]
    \centering
    \includegraphics[width=1\linewidth]{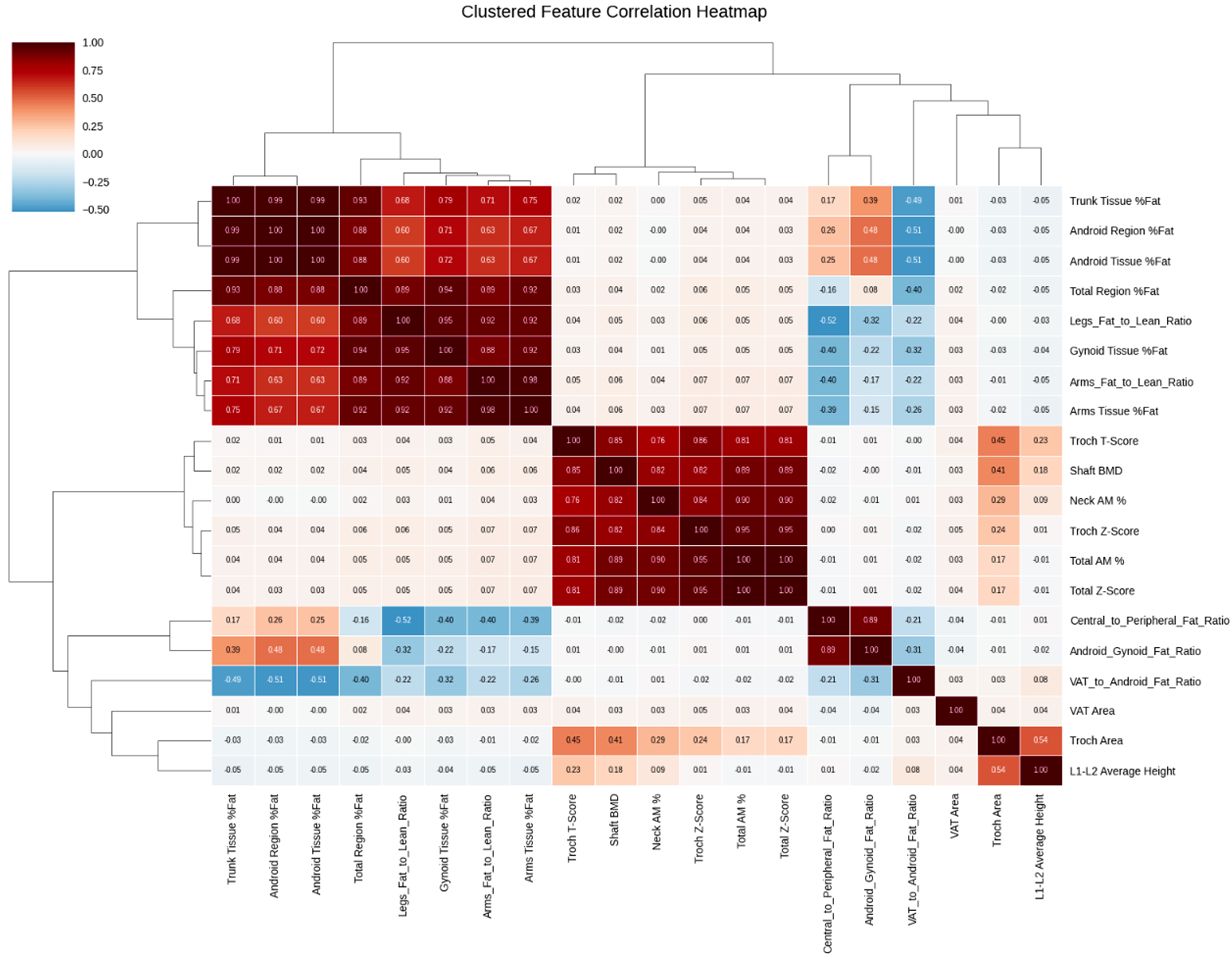}
    \caption{Cluster-based analysis of important features}
    \label{fig:cluster_analysis}
\end{figure*}

\textbf{Ensemble Selection \textemdash}
Features selected if they appear in any of the three methods subject to multicollinearity constraint from Equations \ref{eq:bmd_cv},\ref{eq:pearson_correlation},\ref{eq:mutual_information}. The ensemble/cluster results of the feature integration is shown in Fig \ref{fig:cluster_analysis}. Features are also selected based on percentile thresholds using Eq: (\ref{eq:selection_rule}).

\begin{equation}
\text{Select } X_j \text{ if } \text{Score}_j \geq Q_{100(1-\alpha)}(\{\text{Score}_k\}_{k=1}^p)
\label{eq:selection_rule}
\end{equation}

where $Q_{100(1-\alpha)}$ denotes the $(100(1-\alpha))$-th percentile and $\alpha$ represents the desired selection rate. The final feature map is developed by combining the correlation features, statistical features, and feature importance.

\subsection{Model Evaluation}
This study employed a comparative machine learning approach to evaluate diabetes prediction performance across three distinct algorithmic paradigms: traditional machine learning, large language models, and specialized tabular transformers. All models were trained and evaluated on identical DXA-derived feature sets to ensure fair comparison.

\subsubsection{Traditional ML implementation}
Eight classical machine learning algorithms were implemented as baseline predictors: Logistic Regression with L2 regularization (max\_iter=1000), Random Forest with 100 estimators and maximum depth of 10,  with 100 sequential learners (learning\_rate=0.1, max\_depth=6), SVM using radial basis function kernel with probability estimation, K-Nearest Neighbors (k=5) with distance weighting, Decision Tree with maximum depth of 10, Gaussian Naive Bayes, and AdaBoost with 50 weak learners.

To handle class imbalance, automatic class weighting and 5-fold stratified cross-validation were employed. Top models were optimized via grid search and integrated using soft voting, with ROC-AUC as the primary and F1-score as the secondary selection metric.

\subsubsection{Large Language Model Evaluation}
Three state-of-the-art large language models—Claude 3.5 Sonnet, GPT-4, and Gemini Pro were evaluated for diabetes prediction via standardized API interfaces to ensure uniform testing conditions. Each model was assigned the role of a senior endocrinologist specializing in DXA-based diabetes prediction and prompted using two strategies: few-shot learning and comparative analysis. Model outputs were parsed to extract structured predictions, including risk classification, probability scores (0–1), confidence levels (50–100\%), and clinical reasoning. Probability values were normalized between 0.05 and 0.95, with error handling and fallback mechanisms implemented to ensure consistent and reliable extraction.

\subsubsection{TabTrans}
For this research analysis, a specialized tabular transformer was implemented, incorporating attention mechanisms specifically adapted for structured, non-sequential data. This architecture combines the representational power of transformer models with optimizations designed for tabular healthcare data \cite{huang2020tabtransformer}. Model training utilized the AdamW optimizer with aggressive regularization (learning\_rate=$5\times 10^{-5}$, weight\_decay=$1\times 10^{-5}$) and cosine annealing warm restarts for learning rate scheduling. Gradient clipping (max\_norm=0.5) was applied to prevent training instability. Early stopping was implemented based on minority class recall with patience of 50 epochs, prioritizing sensitivity over overall accuracy. The training process further incorporated an auxiliary minority detection loss (weight=0.3) to enhance model focus on diabetic cases.


\section{Results and Discussions}
The experimental results of this research work are divided into three sections. Firstly, the prediction of diabetes onset using multimodal data and TabTrans model. Secondly, the comparative analysis of TabTrans models with generative AI and typical ML models. Thirdly, the feature interpretation and explaination section that focusing on identifying the risk factors associated with diabetes development in Qatari cohort.

\subsection{Diabetes Prediction}
\begin{figure}[!ht]
    \centering
    \includegraphics[width=1\linewidth]{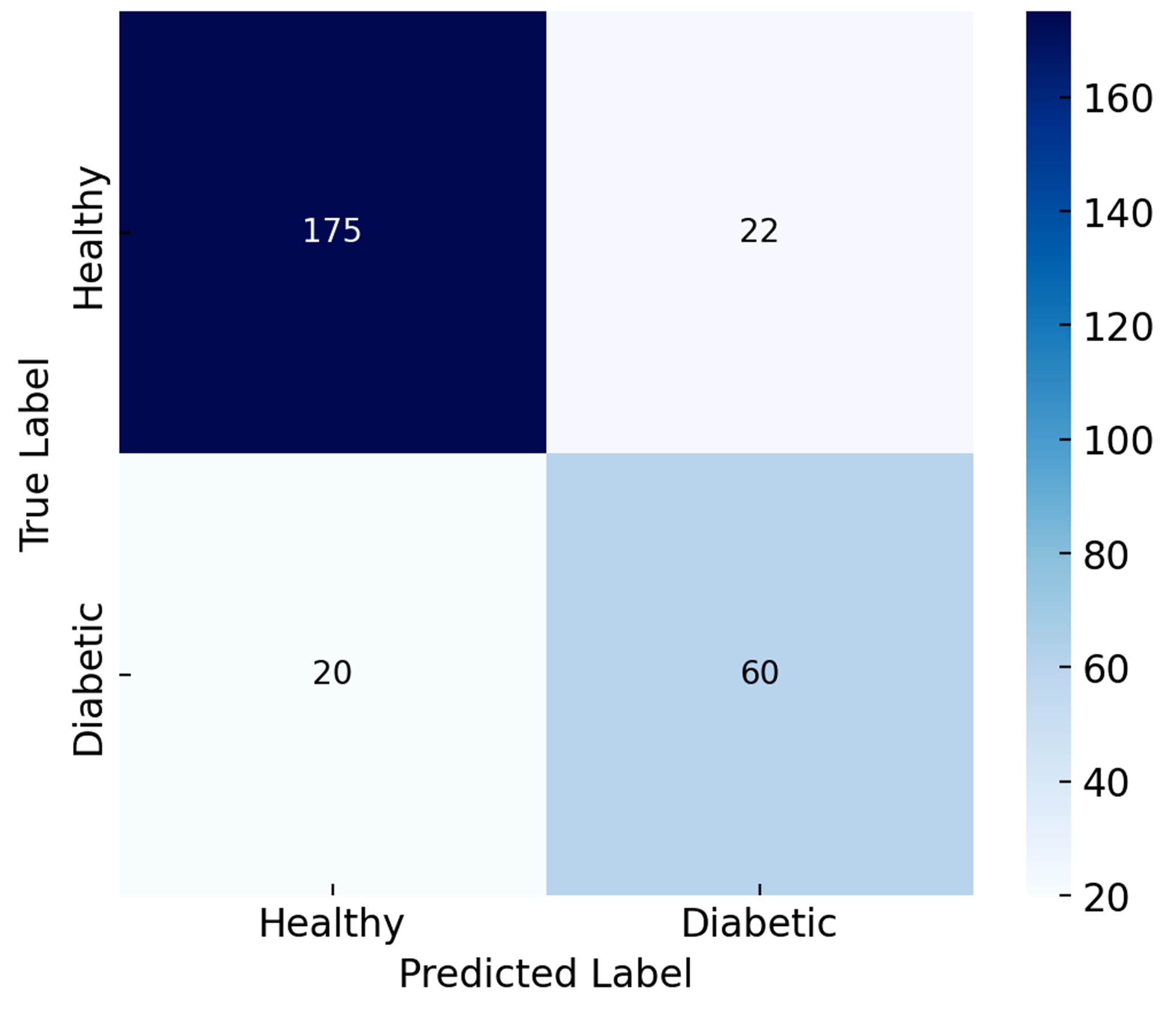}
    \caption{TabTrans performance results for diabetes prediction using multimodal QBB data.}
    \label{fig:conf_matrix}
\end{figure}

After applying the TabTrans model, 175 healthy subject are correctly classified, while 22 misclassified as Diabetic participants. Moreover, 60 diabetic participants are correctly classified, while 20 misclassified as Healthy subjects as shown in Fig.~\ref{fig:conf_matrix}. An overall performance of 84.8\% accuracy, 73.2\% precision, 75.0\% recall, and 74.1\% F1-score is reported for the TabTrans model.

\begin{figure*}[!ht]
    \centering
    \includegraphics[width=1\linewidth]{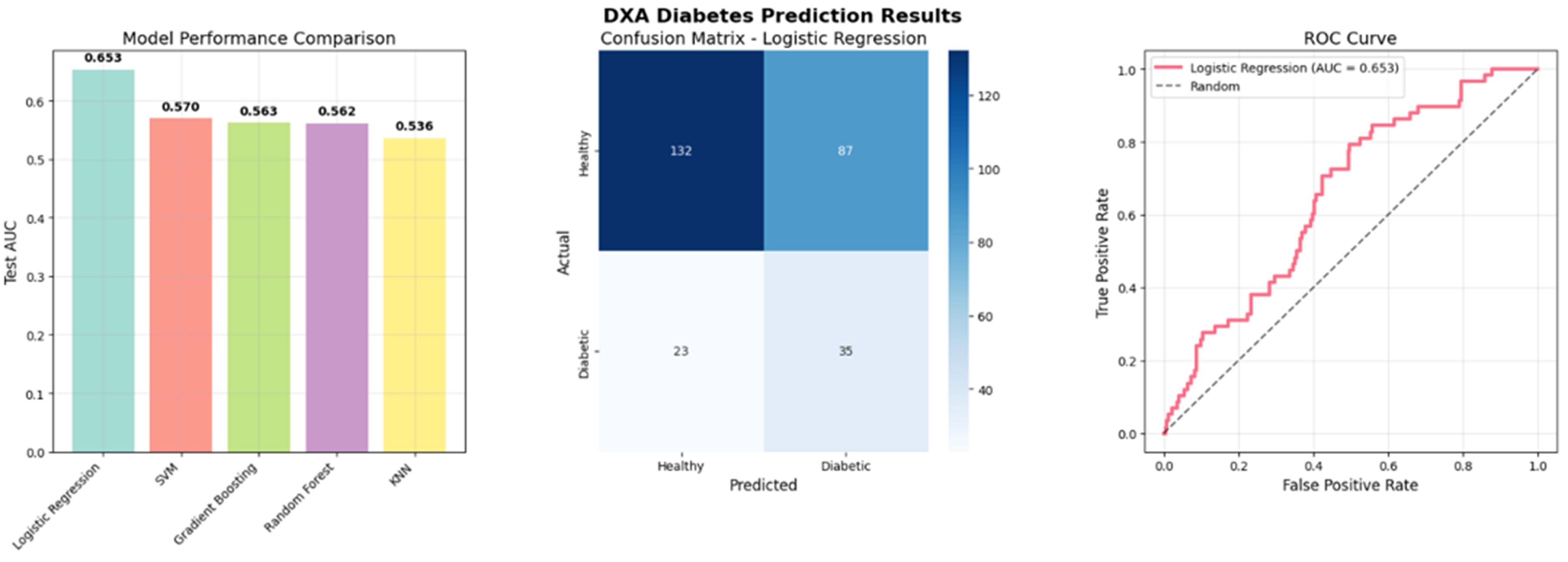}
    \caption{Typical ML performance results for diabetes prediction using multimodal QBB data.}
    \label{fig:conventional}
\end{figure*}

The results of the conventional ML models are shown in Fig. \ref{fig:conventional}. The highest AUC value is reported for linear regression model that 65.3\% much smaller than TabTrans model as shown in Tab. \ref{tab:perf_Res}.

\subsection{Comparative Analysis}
To assess the applicability of the proposed TabTrans model, we compared its performance with both generative AI models and conventional ML models. The results on the same multimodal dataset and feature map are presented in Tab. ~\ref{tab:perf_Res}. The TabTrans model achieved higher performance than the generative AI models and traditional ML models, demonstrating its suitability for tabular data classification and prediction, particularly with small datasets. The dataset used was limited and imbalanced, comprising 1,103 healthy and 279 diabetic participants. After training–testing splits, the reduced sample size led to overfitting and higher misclassification rates in other conventional ML models.

\begin{table*}[!ht]
\centering
\caption{\label{tab:perf_Res}Performance analysis of TabTran vs generative models and conventional ML models.}
\begin{tabular}{|c|c|c|c|c|c|}
\hline
    \textbf{Model Name} & \textbf{Accuracy (\%)} & \textbf{Precision (\%)} & \textbf{Recall (\%)} & \textbf{ROC AUC (\%)} & \textbf{Hyper-parameter Tuning (AUC \%)}  \\ \hline\textbf{}
    Claude 3.5 Sonnet & 70.3 & 66.5 & 67.8 & 74.7 & \textemdash \\ \hline
    GPT-4 & 56.7 & 51.8 & 53.1 & 60.6 & \textemdash \\ \hline
    Gemini Pro & 61.2 & 58.4 & 56.7 & 71.3 & \textemdash \\ \hline
    Logistic Regression & 57.8 & 25.6 & 53.4 & 61.7 & 65.3 \\ \hline
    Support Vector Machine & 54.3 & 48.2 & 54.7 & 56.2 & 57.0 \\ \hline
    K-Nearest Neighbor & 46.2 & 41.8 & 47.1 & 46.4 & 53.6 \\ \hline
    Decision Tree & 51.5 & 48.3 &  50.2 & 44.3 & 56.6 \\ \hline
    Gaussian Naive Bayes & 48.6 & 41.3 & 47.2 & 45.7 & 52.3\\ \hline
    AdaBoost & 51.3 & 47.5 & 49.3 & 48.7 & 56.3 \\ \hline
    Random Forest & 49.2 & 46.2 & 47.9 & 49.8 & 56.2\\ \hline
    \textbf{Tabular Transformers} & \textbf{84.8 } & \textbf{73.2} & \textbf{75} & \textbf{79.7} & \textbf{59.1} \\ \hline
\end{tabular}
\end{table*}

\subsection{Feature Interpretation}
\begin{figure*}[!ht]
    \centering
    \includegraphics[width=0.87\linewidth]{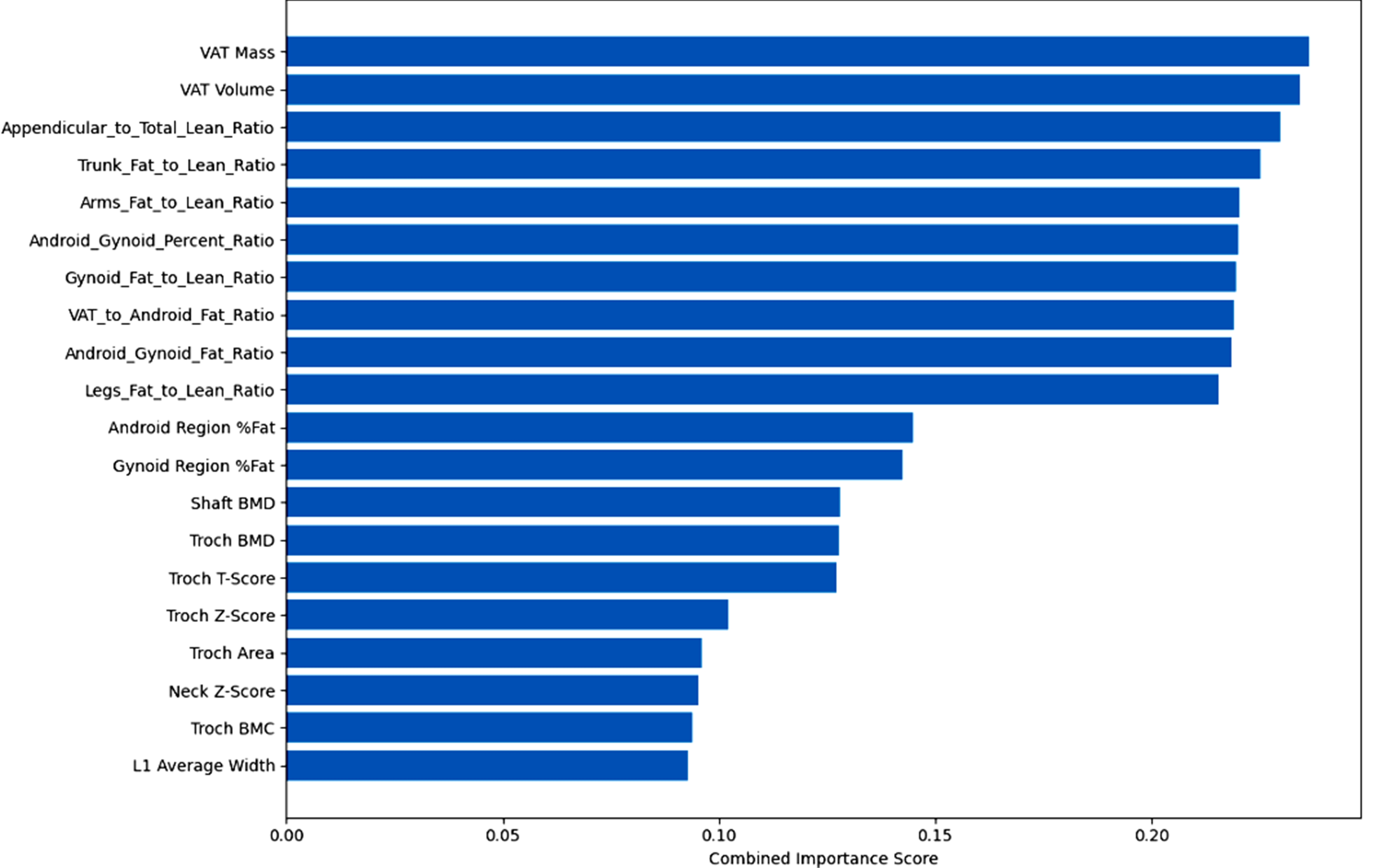}
    \caption{Feature interpretation to identify DXA-based features associated with diabetes onset in QBB cohort.}
    \label{fig:interpretation}
\end{figure*}

To perform feature interpretation, we used probabilistic model. The results are shown in Fig. ~\ref{fig:interpretation}, where X-axis represents how much each feature contributes to the model’s predictions. The models classified the feature importance into three sections including highly contributed features, moderate contributors, and least important features.

\begin{itemize}

    \item \textbf{Highly Contributors \textemdash} Higher values mean the feature is more influential in classification/regression. While Y-axis represents Features (biological/clinical measures). From Fig. ~\ref{fig:interpretation}, it is observed that VAT Mass and VAT volume are the most important predictors. This means visceral fat is strongly linked to the outcome your model is predicting (likely diabetes, metabolic risk, or another health condition). Appendicular-to-Total Lean Ratio, Trunk Fat-to-Lean Ratio, arms Fat-to-Lean Ratio, Legs Fat-to-Lean Ratio are also highly ranked. These reflect regional fat vs. muscle balance, which is clinically relevant for metabolic health. Features like Android Gynoid Percent Ratio, VAT to Android Fat Ratio, and  Android Gynoid Fat Ratio capture central vs. peripheral fat distribution. Higher android (abdominal) fat compared to gynoid (hip/thigh) fat is a known risk marker.

    \item \textbf{Moderate Contributors \textemdash} BMD and Z-scores/T-scores of skeletal regions (shaft, troch, neck) appear lower on the list, suggesting weaker predictive power relative to fat measures.
    
    \item \textbf{Least Important Feature \textemdash} L1 Average Width is at the bottom, meaning it had the smallest contribution.
\end{itemize}

The proposed TabTrans finds that VAT and fat distribution ratios are much more predictive of the diabetes prediction condition (likely diabetes or metabolic risk) than bone-related features.

T2DM exhibits two fundamental metabolic features: excess body weight and hyperinsulinemia. These pathophysiological traits are established contributors that demonstrate beneficial impacts on BMD and additional skeletal metrics. Importantly, statistical adjustment for BMI in clinical investigations reveals that the diabetes-bone density correlation persists. Similarly, the interplay between excess weight and skeletal density in diabetic populations presents considerable complexity. While DXA measurements in overweight diabetic subjects may encounter modest technical limitations, these variations remain minimal and can produce either increased or decreased values without substantially compromising diagnostic precision. Fat tissue releases numerous signaling proteins termed adipokines, which modulate bone turnover mechanisms through both direct and indirect pathways. Resultantly, diabetic males frequently demonstrate elevated leptin concentrations compared to their non-diabetic counterparts. Leptin functions to suppress bone breakdown by reducing RANK/RANKL signaling while concurrently promoting osteoprotegerin production. Other adipose-derived factors, including adiponectin and resistin, are similarly present within bone-building (osteoblast) and bone-degrading (osteoclast) populations. Though the specific effects of these signaling molecules on skeletal metabolism require further clarification, current research indicates their involvement in guiding mesenchymal stem cell commitment toward either osteogenic or adipogenic developmental pathways \cite{umur2024exploring, rajput2024osteogenic}.

\section{Limitations}
Despite its promising outcomes, this study has several limitations. The analysis is confined to Qatari participants from the Qatar Biobank, limiting generalizability to other populations due to genetic and cultural differences. The modest sample size (1,382 participants) and class imbalance raise concerns about model generalization, while limited longitudinal data only 45 subjects with eight-year clinical records restricts long-term risk assessment. Reliance on self-reported information may introduce measurement bias, and the exclusion of intermediate glucose tolerance, monogenic, and autoimmune diabetes cases narrows clinical scope. Nonetheless, the study provides a valuable framework for minimally invasive diabetes prediction using DXA-based analysis.

\section{Conclusion}
This study presents a novel methodology for predicting diabetes development over time using transformers and multimodal data derived from DXA scans and electronic health records. By integrating bone-related measurements from multiple anatomical sites (femur, spine, total body scans, etc.) and EHR data (LDL, HDL, HbA1c, demographic data, etc.), we implemented tabular transformers, generative AI, and conventional machine learning algorithms. Our experimental results demonstrated that TabTrans outperformed generative AI tools, but underperformed conventional ML models. Through probabilistic interpretation methods, we identified key risk indicators (Vat mass, Vat volume, Ward BMD, Ward BMC, Torch T-scores and Z-scores, Trunk and Arm Lean ratio, L1 - L4 scores) associated with diabetes development in Qatari adults.

The robust performance results demonstrate the promising clinical utility of fast, non-invasive diagnostic methods in healthcare settings. Our findings suggest that machine learning approaches can effectively derive meaningful diagnostic information from sophisticated medical imaging data, opening pathways for enhanced early disease detection. However, extensive clinical trials are needed to confirm the diagnostic associations with diabetes development.

\section*{Acknowledgment}
The authors acknowledge Qatar Biobank (Project QF-QBB-RES-ACC-00181) for providing access to the data, and thank the laboratory and spectrometry teams for their assistance by providing the clinical and bones data. We also thanks Morena M. Altura and Elizabeth Jose for their valuable support and guidance during the accumulation of QBB data. The authors declare no conflict of interest.


\section*{Concent for Publication}
The authors utilized the ChatGPT tool to assist in collecting background information for this study. All research content has been thoroughly reviewed and edited by authors, who take full responsibility for the accuracy and integrity of the final study.


{
    \bibliographystyle{ieeetr}
    \bibliography{ref}
}

\end{document}